\documentclass[11pt,twoside]{article}
\usepackage{asp2014}

\aspSuppressVolSlug
\resetcounters

\begin{document}

\title{Learning the Night Sky with Deep Generative Priors}

\author{Fausto~Navarro, Daniel~Hall, Tam\'{a}s~Budav\'{a}ri, and Yashil~Sukurdeep}
\affil{Johns Hopkins University, Baltimore, Maryland, United States}
\paperauthor{Fausto~Navarro}{fnavarr3@jhu.edu}{ORCID}{}{Applied Mathematics and Statistics}{Baltimore}{Maryland}{21218}{United States}
\paperauthor{Daniel~Hall}{deh275@cornell.edu}{ORCID}{}{Applied Mathematics and Statistics}{Baltimore}{MD}{21218}{United States}
\paperauthor{Tamas~Budav\'{a}ri}{budavari@jhu.edu}{ORCID}{}{Applied Mathematics and Statistics}{Baltimore}{MD}{21218}{United States}
\paperauthor{Yashil~Sukurdeep}{yashil.sukurdeep@jhu.edu}{ORCID}{}{Applied Mathematics and Statistics}{Baltimore}{MD}{21218}{United States}
  
\begin{abstract}
Recovering sharper images from blurred observations, referred to as \textit{deconvolution}, is an ill-posed problem where classical approaches often produce unsatisfactory results. In ground-based astronomy, combining multiple exposures to achieve images with higher signal-to-noise ratios is complicated by the variation of point-spread functions across exposures due to atmospheric effects. We develop an unsupervised multi-frame method for denoising, deblurring, and coadding images inspired by deep generative priors. We use a carefully chosen convolutional neural network architecture that combines information from multiple observations, regularizes the joint likelihood over these observations, and allows us to impose desired constraints, such as non-negativity of pixel values in the sharp, restored image. With an eye towards the Rubin Observatory, we analyze 4K by 4K Hyper Suprime-Cam exposures and obtain preliminary results which yield promising restored images and extracted source lists.
\end{abstract}

\section{Introduction}
\label{sec:intro}
The latest generation of ground-based astronomical surveys aim to capture exposures of large swathes of the night sky to advance our understanding of astronomy~\citet{ivezic2019lsst}. Processing pipelines for these surveys will have to address the presence of unwanted atmospheric blur. \textit{Deconvolution}, the process of removing this blur, is complicated by the high level of noise in the exposures, their high dynamic range, and the presence of artifacts and obstructions in the image. We tackle the specific problem of multi-frame astronomical image reconstruction, which entails combining multiple blurry, ground-based astronomical exposures in order to produce a single, sharp image of the night sky. Previous approaches to address the problem include lucky imaging~\citet{tubbs2003lucky}, coadding~\citet{annis2014sloan}, maximum likelihood estimation~\citet{schulz1993multiframe, zhulina2006multiframe}, and streaming methods~\citet{harmeling2009online, harmeling2010multiframe, hirsch2011online, lee2017robust, lee2017streaming}. We develop a novel unsupervised, multi-frame method inspired by deep generative priors, outlined in Section~\ref{sec:model}. 

\section{Model and Approach}
\label{sec:model}
We begin by describing the model for our approach. Given a sample of observations $y\!\equiv\!\{y^{(1)}, y^{(2)}, \dots, y^{(n)}\}$, we model each observation $y^{(t)}$ as the convolution of a common latent image $x$ with a point-spread function (PSF) $f^{(t)}$, plus an additive error term $\eta^{(t)}$. We highlight that the PSFs and error terms can vary from exposure to exposure. While photon counts in the raw exposures follow a Poisson distribution, the large number of photons allows us to model the sky-subtracted images as a Gaussian with zero mean and variances $v^{(t)}$, which are usually given to the user in astronomical imaging pipelines. Thus, our model for each pixel value in each exposure, denoted $y^{(t)}_{ij}$, is
\begin{equation}
    \label{eq:model_exposures}
    y^{(t)}_{ij} = \left[f^{(t)} \!* x\right]_{ij} + \eta^{(t)}_{ij}, \quad \textrm{where}\quad \eta^{(t)}_{i j} \sim N\left(0, v^{(t)}_{ij}\right) .
\end{equation}

One could attempt to find the latent image $x$ (and the PSFs $f\!\equiv\!\{f^{(1)}, f^{(2)}, \dots, f^{(n)}\}$ if they are unknown) as \textit{maximum likelihood estimates} (MLE) of the model above. However, such methods often fail to form meaningful estimates~\citet{schulz1993multiframe, zhulina2006multiframe}. We thus operate under a Bayesian framework, and solve for $x$ and $f$ as \textit{maximum a posteriori} (MAP) estimates
\begin{equation}
    \label{eq:map_estimate}
    \Hat{x}, \Hat{f} = \underset{x, f}{\mathrm{argmax}} \ \ln p(y \mid x, f) + \ln p(x, f) .
\end{equation}
Note that $p(y \mid x, f)$ is the conditional distribution of the exposures $y$ given the latent image $x$ and PSFs $f$, which is the Gaussian distribution from~\eqref{eq:model_exposures}. Meanwhile, $p(x, f)$ is a prior distribution on the latent image and PSFs, for which a handcrafted regularization prior such as the total variation norm might traditionally be used. However, one can impose an effective regularizing prior through the structure of an untrained, generative neural network, i.e., a so-called \textit{deep generative prior}~\citet{ulyanov2018deep}. 

\begin{figure}[ht]
    \centering   
    \includegraphics[width=0.95\linewidth]{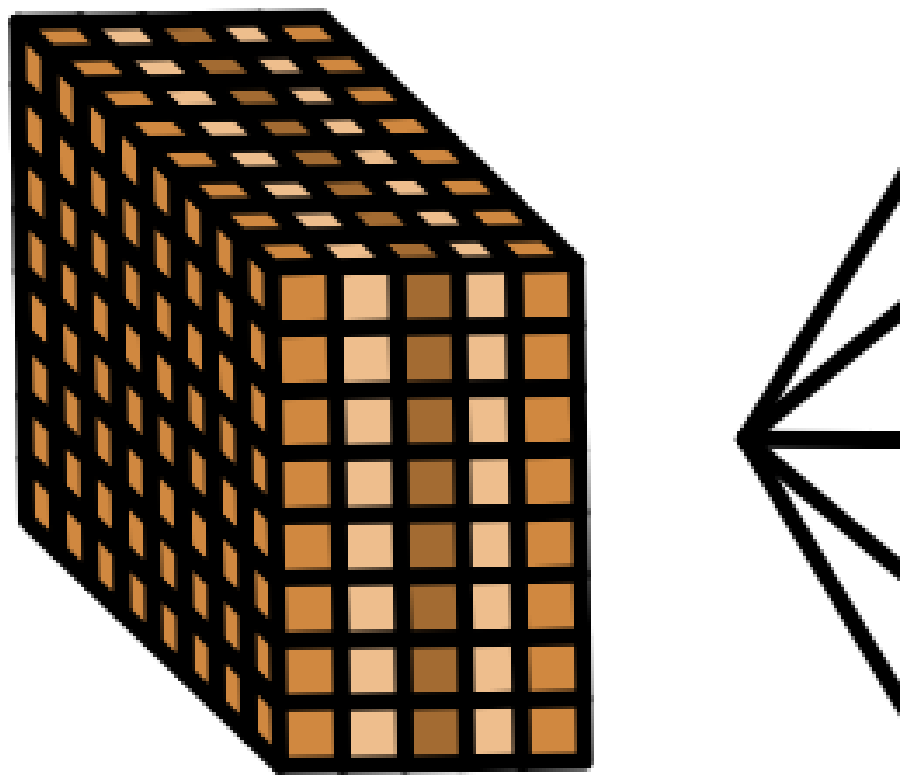}
    \caption{Network architecture. We extract and combine multi-scale information from the exposures via various convolutional layers. All input exposures first pass through several depth-wise convolution layers in parallel. The output channels are concatenated, and a pointwise convolution completes the ``encoder" part of our network to produce the latent image $x$. Desired constraints, such as non-negativity of pixel values in $x$, are enforced by applying a ReLU activation.  We then ``decode" $x$ via a final 2D convolutional layer in order to produce the reconstructions.}
    \label{fig:network_architecture}
\end{figure}

Inspired by this approach, we develop an unsupervised multi-frame method for deconvolving astronomical images. Our approach is an extension of the flash-no flash method for image-pair restoration in~\citet{ulyanov2018deep}, to the setting of multi-frame image reconstruction. In our framework, we encode the latent image $x$ as a \textit{function} of the multiple exposures $y\!\equiv\!\{y^{(1)}, y^{(2)}, \dots, y^{(n)}\}$. We parametrize this function via a neural network with learnable parameters $\theta$ and denote it by $F_{\theta}$. We then decode the latent image $x = F_{\theta}(y)$ by convolving it with $n$ convolutional filters $f\!\equiv\!\ \{f^{(1)}, f^{(2)}, \dots, f^{(n)}\}$ in order to produce reconstructions of our input exposures, denoted by $\Hat{y}\!\equiv\!\{\Hat{y}^{(1)}, \Hat{y}^{(2)}, \dots, \Hat{y}^{(n)}\}$ where $\Hat{y}^{(t)} = f^{(t)} * F_{\theta}(y)$. The convolutional filters $f$ could be the PSFs corresponding to each exposure if these are known, otherwise they could be additional learnable parameters of the network. We refer the reader to Figure~\ref{fig:network_architecture} for additional details about the network's architecture.

To tune the parameters of our network, we minimize the Huber loss between our network's inputs and outputs (scaled by their corresponding standard deviations), i.e.,
\begin{equation}
    \label{eq:loss_network}
     \theta^* = \underset{\theta}{\mathrm{argmin}} \sum_{t=1}^n \sum_{i,j} L_{\delta} \left(\frac{y_{ij}^{(t)}}{\sqrt{v_{ij}^{(t)}}}, \frac{\left[ f^{(t)} * F_\theta(y) \right]_{ij}}{\sqrt{v_{ij}^{(t)}}}\right) ,
\end{equation}
where the Huber loss is applied pixel-wise and is defined as
\begin{equation}
    \label{eq:huber_loss}
    L_{\delta}(y, \Hat{y}) \coloneqq \begin{cases}
        \frac{1}{2} \left(y - \Hat{y} \right)^2  & \text{ for } | y - \hat{y} | \leq \delta , \\
        \delta \left( | y - \Hat{y} | - \frac{1}{2} \delta \right)  & \text{ otherwise. }
    \end{cases}
\end{equation}

We typically set $\delta = 1$ in our experiments. Note that the Huber loss behaves like the mean squared error but is more resistant to outliers, which makes our recovered latent image $\Hat{x}$ and reconstructions $\Hat{y}$ robust to heavy-tailed noise or saturated pixels in the exposures. For emphasis, we highlight that the deconvolved latent image $\Hat{x}$ is computed via a forward pass through the trained encoder part of our network, i.e.,
\begin{equation}
    \label{eq:latent_image_forward_pass}
    \Hat{x} = F_{\theta^*}(y). 
\end{equation}

\section{Results}
\label{sec:results}
\begin{figure}[ht]
    \centering
    \includegraphics[width=0.32\linewidth]{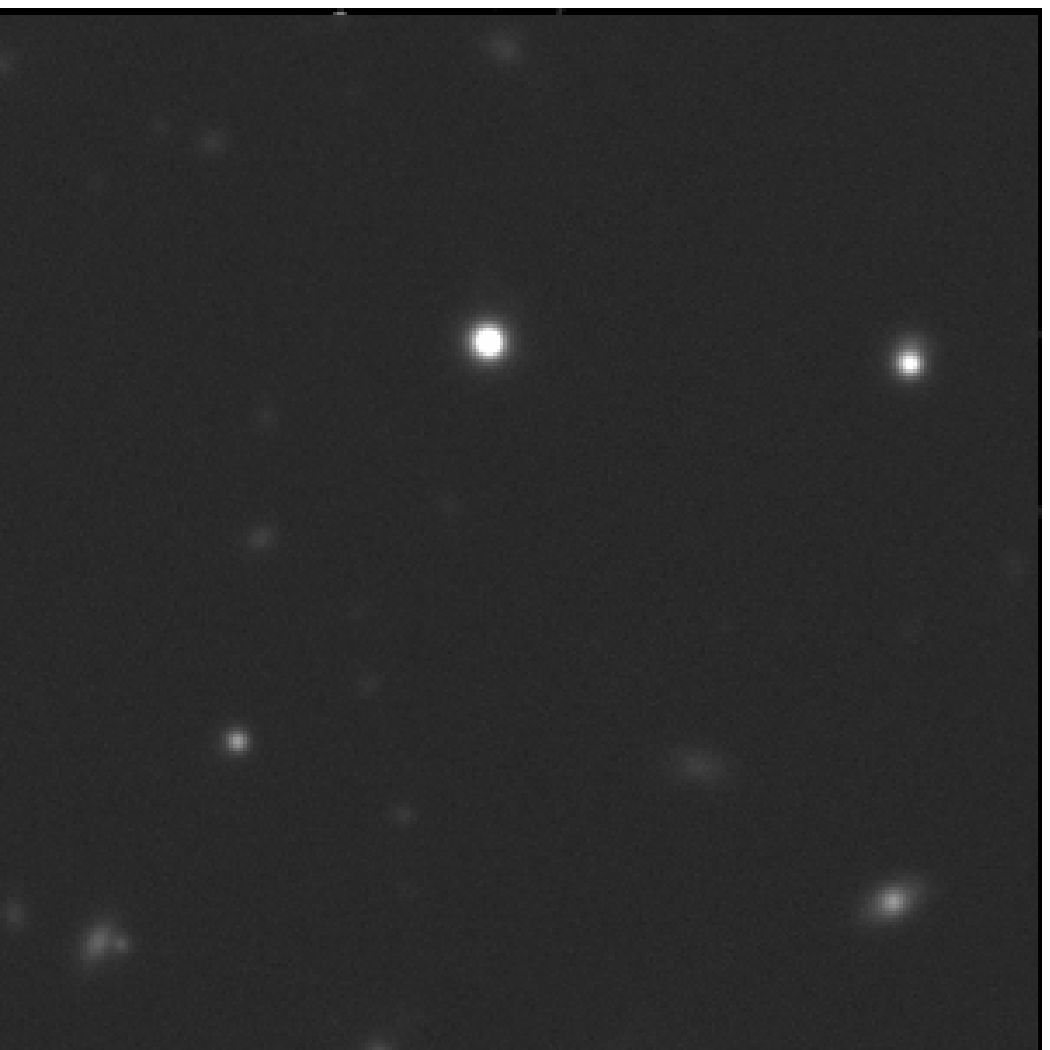}
    \includegraphics[width=0.32\linewidth]{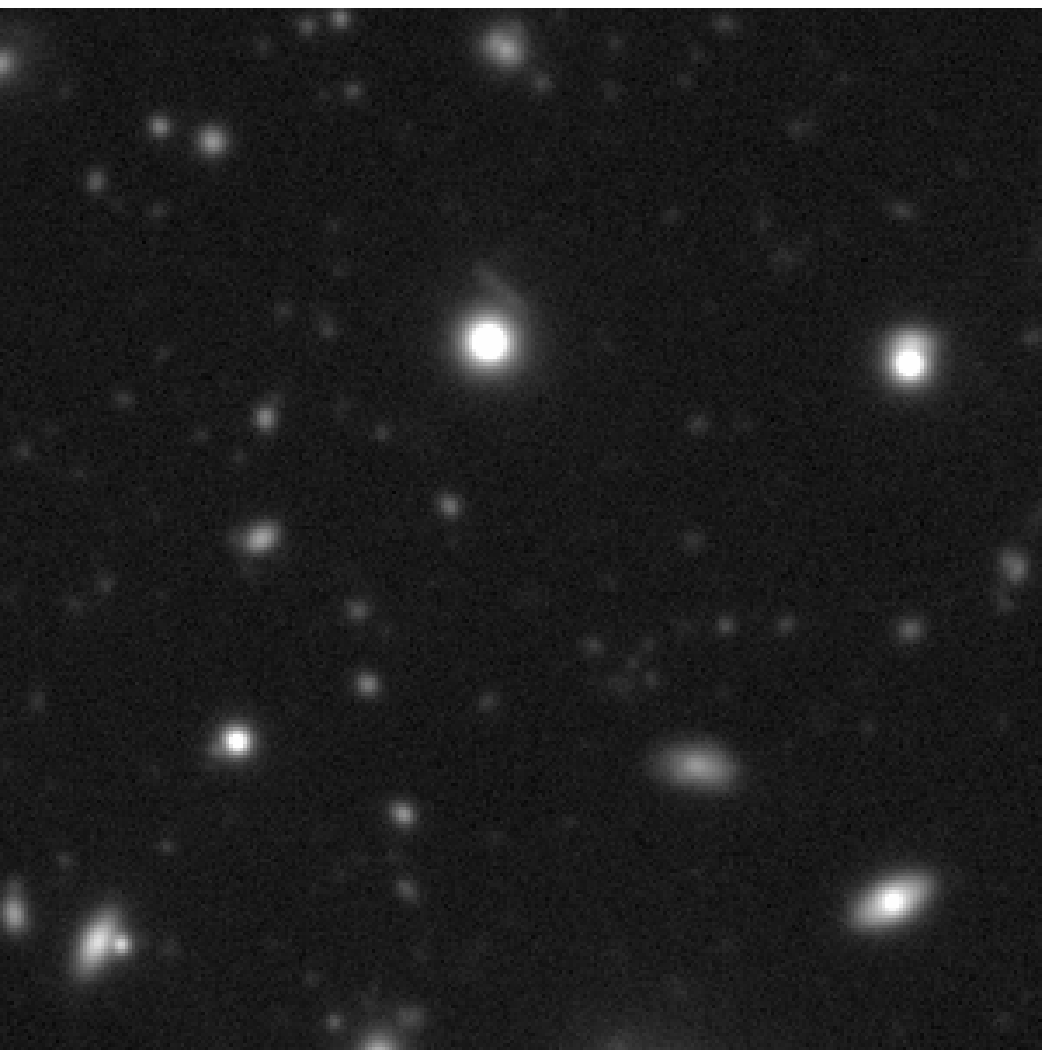}
    \includegraphics[width=0.32\linewidth]{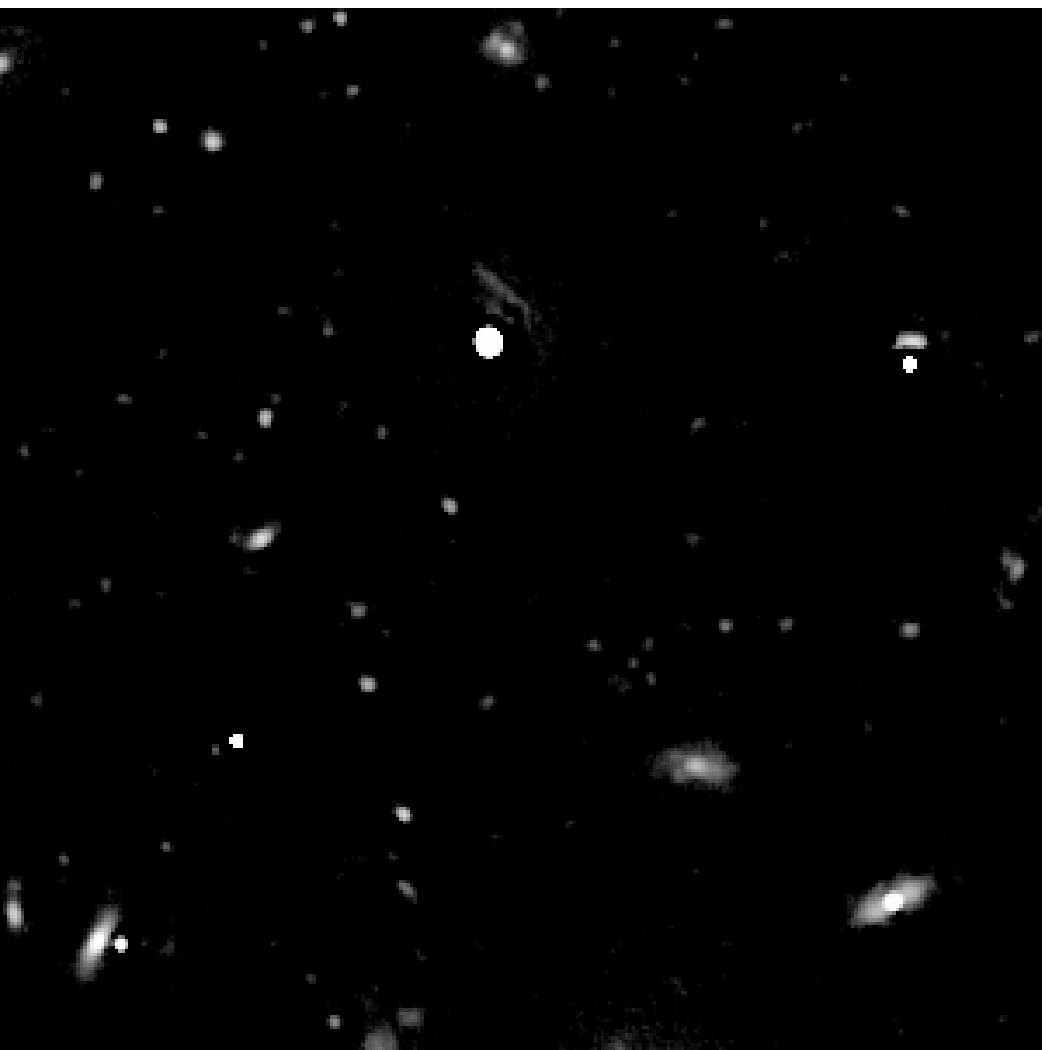}
    \includegraphics[width=0.32\linewidth]{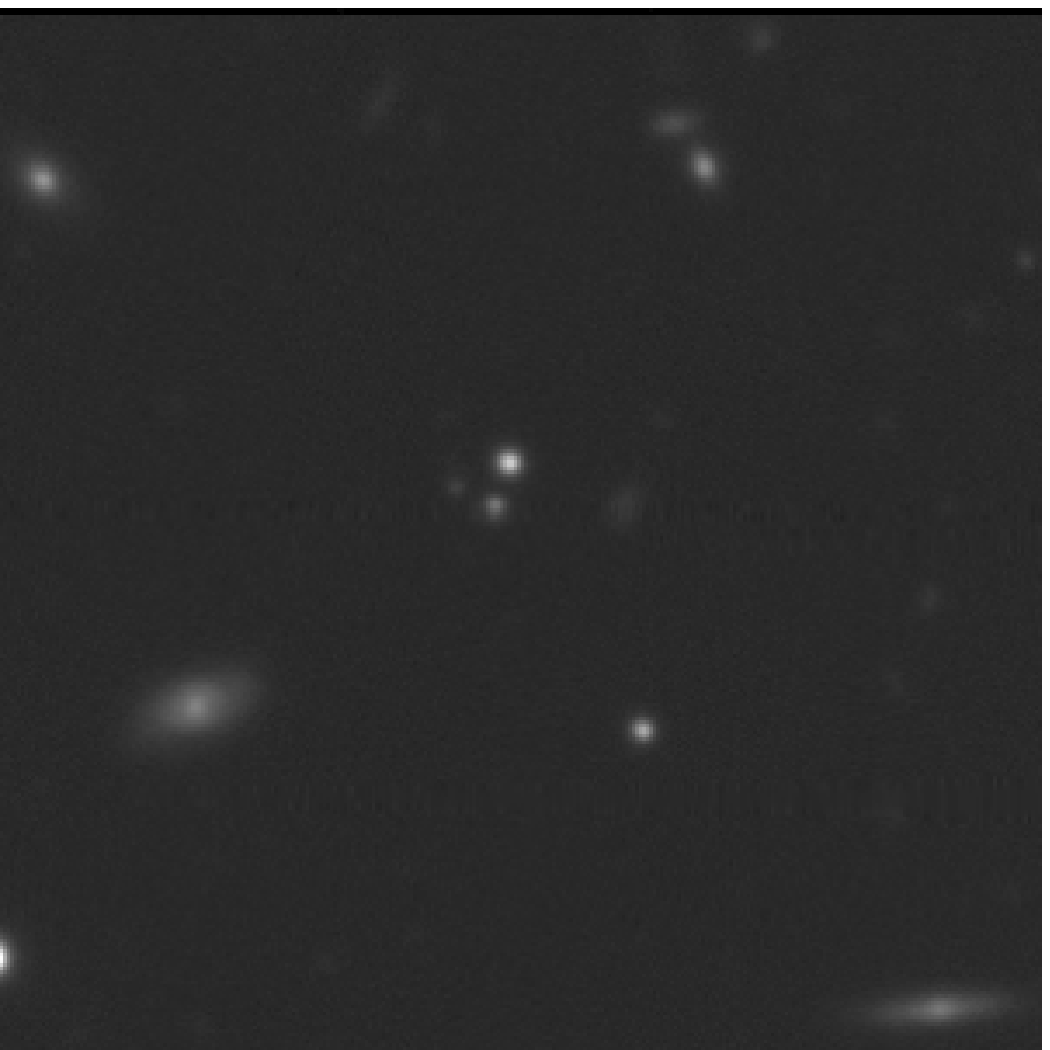}
    \includegraphics[width=0.32\linewidth]{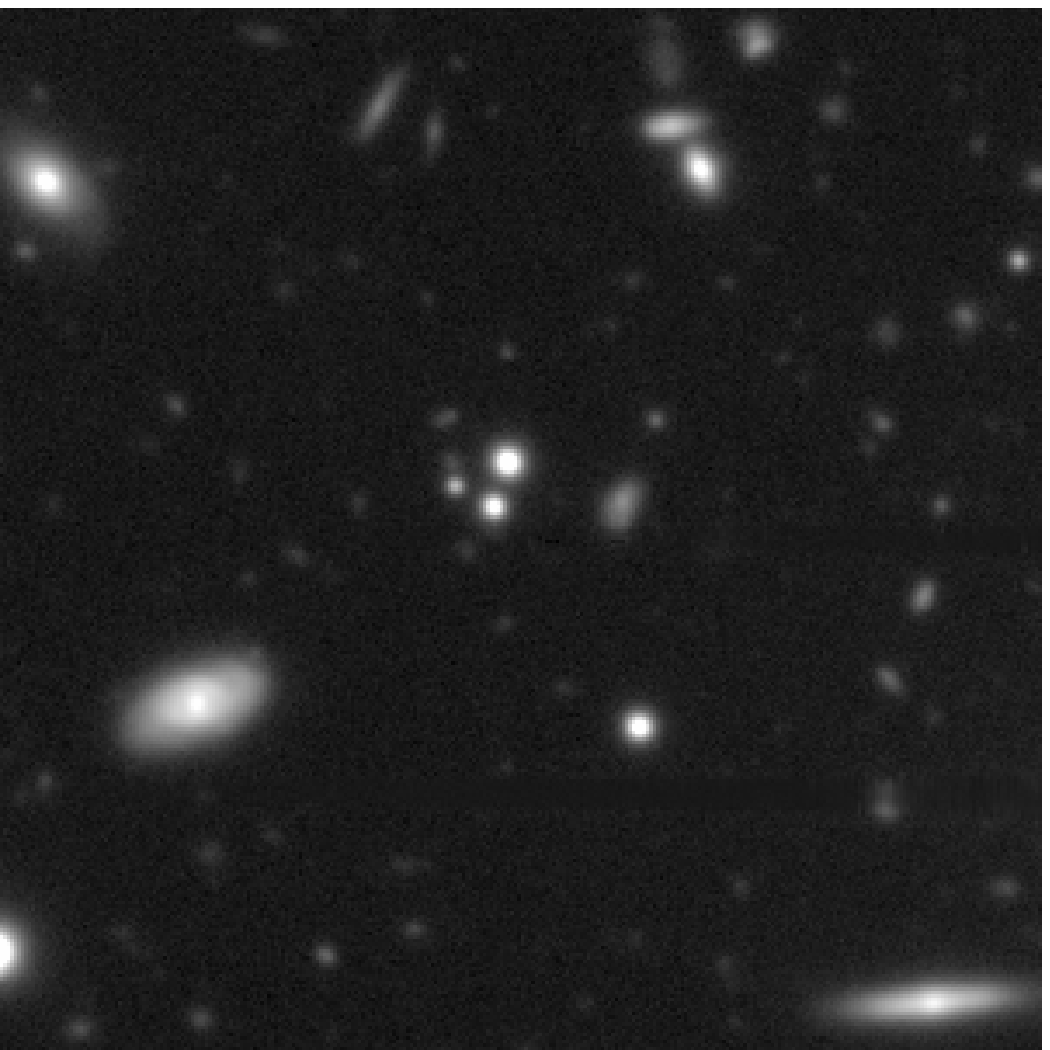}
    \includegraphics[width=0.32\linewidth]{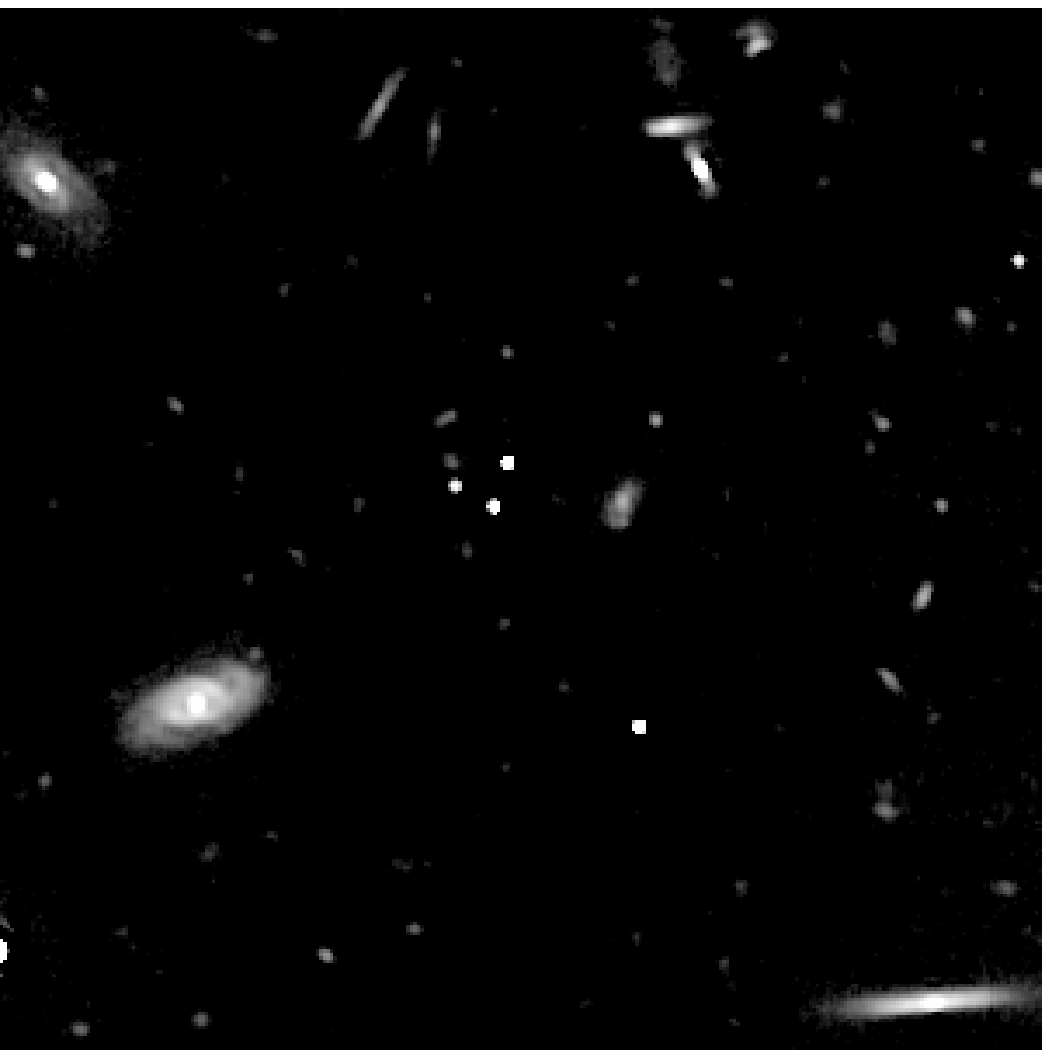}
    \caption{Comparison: Selected cutouts from a raw exposure (left) vs. the ``sample mean" co-add (middle) vs. a restored image $\Hat{x}$ from our approach (right). Our method produces a physically meaningful restored latent image of the night sky which is suitable for photometry. Pixels in the sky background have zero intensity, with only minor speckles of noise appearing around bright objects, unlike the co-add where noise is uniformly present across the image. Moreover, our method successfully deblurs a wide array of sources, resulting in e.g. galaxy shapes being visibly more well-defined, and large blurry stars appearing as well resolved point sources.}
    \label{fig:comparison_nn_coadd}
\end{figure}

We tested our method on a set of $33$ exposures from the Hyper Suprime-Cam telescope, which are closest in quality to imaging data from the upcoming Rubin Observatory. We compare the latent image $\Hat{x}$ obtained using our approach with a ``naive" co-add of the exposures, calculated by taking their sample mean. Results in Figure~\ref{fig:comparison_nn_coadd} demonstrate a significant improvement in the quality of the reconstruction obtained via our approach.  

\section{Conclusion and Future Work}
\label{sec:conclusion}
We have introduced a novel method for multiframe astronomical image deconvolution based on deep generative priors. The key to our method lies in encoding the latent image of the sky as a function of the multiple observed ground-based exposures, and parametrizing this function via a convolutional neural network. Preliminary results on imaging data from the Hyper Suprime-Cam telescope yield physically meaningful restorations that are suitable for photometry. As future work, we plan to extend our approach to perform image reconstruction with observations from several color or frequency bands. Another natural extension involves adapting our model so that it learns a super-resolved latent image in which one obtains sub-pixel detail in galaxies and stars, thus enabling improved photometry.

\bibliography{P11}

\end{document}